\documentclass[letterpaper, 10 pt, conference]{ieeeconf}

\IEEEoverridecommandlockouts                             

\usepackage{graphicx}
\usepackage{multicol}
\usepackage{booktabs}
\usepackage{amsmath,amssymb}
\usepackage[utf8]{inputenc}
\usepackage[english]{babel}
\usepackage{multirow}
\usepackage[font=small]{caption}
\usepackage{float}
\usepackage[hidelinks]{hyperref}

\usepackage{lettrine}
\usepackage[]{algorithm2e}
\usepackage{algpseudocode}
\usepackage{cite}
\usepackage{filecontents}
\usepackage{lipsum}
\usepackage{color}
\usepackage{esdiff}
\usepackage{epstopdf}
\usepackage[normalem]{ulem}
\usepackage{soul}

\usepackage{subcaption}
\usepackage{xcolor}
\usepackage{colortbl}
\usepackage{balance}
\definecolor{pink}{rgb}{1, 0, 1}
\definecolor{orange}{rgb}{1, 0.7529, 0}
\definecolor{darkgreen}{rgb}{0, 0.8, 0}
\begin{document}
\bstctlcite{IEEEexample:BSTcontrol}

\title{A Hierarchical Coverage Path Planning Algorithm for Unknown Environments}

\author{Zongyuan Shen$^1$, Haodong Liu$^1$, Gao Wang$^1$, Hongbin Ma$^2$, Yaming Ou$^3$, \\ Shancheng Zhao$^1$, and Dehua Zhou$^1$
\thanks {$^1$College of Information Science and Technology, Jinan University, Guangzhou 510632, China.}
\thanks{$^2$School of Automation, Beijing Institute of Technology, Beijing 100081, China.}
\thanks {$^3$School of Artificial Intelligence, University of Chinese Academy of Sciences, Beijing 100049, China.}
}
        
\maketitle

\begin{abstract}
This paper presents an online coverage path planning algorithm for unknown environments. During navigation, the initially unknown search area is progressively decomposed into disconnected subareas as new obstacle information is acquired and coverage proceeds. These subareas are organized in an incrementally constructed decomposition tree that preserves their hierarchical parent-child relationships. Based on this tree, a global coverage tour is maintained and updated online by prioritizing newly generated child subareas according to their exploration states and distances from the robot. A local planner then generates coverage motions within each selected subarea, allowing the robot to adapt its trajectory as the environment is gradually revealed. Its performance is evaluated via high-fidelity simulations in complex scenarios. The results show improved coverage efficiency in terms of path length and overlap ratio in comparison to three baseline algorithms.

\end{abstract}

\section{Introduction}

Recent advances in sensing, computing, and autonomous systems have enabled robots to perform an increasing variety of service and field operations. Many of these tasks require a robot to traverse a target workspace and provide complete coverage of all accessible regions, giving rise to the coverage path planning (CPP) problem. CPP is a fundamental problem in robot motion planning and has been widely applied to environmental monitoring~\cite{shen2016autonomous,shen2017,Ou2025}, structural inspection~\cite{vidal2017,veeraraghavan2024complete}, surface cleaning~\cite{yang2020cellular}, spray painting~\cite{vempati2018paintcopter}, and agricultural operations~\cite{maini2022online}. In addition to guaranteeing complete coverage, practical CPP methods are expected to achieve efficient execution by reducing path length, repeated coverage, the number of turns, and other task-dependent costs.

In many practical applications, however, the environment is unknown or only partially known before deployment. In such cases, a complete coverage path cannot be planned in advance, and the robot must incrementally generate and update its coverage trajectory using environmental information collected by onboard sensors. This online setting couples coverage with exploration, requiring the robot to discover previously unknown regions while simultaneously avoiding obstacles and covering the observed free space. The planning problem becomes particularly challenging in complex environments, where inappropriate coverage decisions may lead to dead ends, isolated uncovered regions, repeated traversal, and excessive overlap. Therefore, an effective online CPP method must not only adapt to newly acquired environmental information in real time, but also make coverage decisions that maintain high efficiency throughout the entire operation.

Most existing online CPP methods determine the next coverage motion primarily from locally available information. Such local decision rules are computationally efficient and easy to implement in real time, but they provide only a limited view of the remaining coverage task. In complex environments, locally favorable decisions may lead the robot toward dead ends, leave isolated uncovered regions, or require repeated traversal to revisit previously bypassed areas. Some methods attempt to alleviate these problems by checking whether a candidate motion would disconnect the remaining uncovered space, but such decisions are still made within a limited local neighborhood and may not capture the broader structure of the coverage task. 

In this regard, we propose a hierarchical coverage planning method for unknown environments that combines online space decomposition, global tour planning, and local coverage planning. During online operation, the search space is progressively decomposed as newly observed obstacles and covered regions separate the remaining uncovered space into multiple subareas. Instead of treating these subareas as independent coverage tasks, the proposed method organizes them into an incrementally constructed decomposition tree that preserves their parent-child relationships. Based on this tree, the global planner continuously updates the visitation sequence of the active subareas by prioritizing newly generated descendants of the current branch. The child subareas are further ordered according to their exploration states and distances from the robot, allowing the planner to coordinate global coverage while retaining a lightweight online planning structure. A local planner then generates the coverage motion within the selected target subarea.

The main contributions of this work are twofold. First, we develop an online space decomposition framework that incrementally organizes dynamically generated uncovered subareas into a hierarchical decomposition tree, preserving the parent-child relationships revealed during coverage. Second, we propose a lightweight global tour planning strategy that directly exploits this tree structure to update the visitation sequence of active subareas using exploration-state and distance-based priorities, while local coverage is performed within the selected target subarea. The effectiveness of the proposed method is evaluated through simulation and real-world experiments. The remainder of this paper is organized as follows. Section~\ref{sec:review} reviews related work on online coverage planning. Section~\ref{sec:algorithm} presents the proposed algorithm, including online space decomposition, global tour planning, and local coverage planning. Section~\ref{sec:results} reports the simulation results, and Section~\ref{sec:conclusions} concludes the paper.

\section{Related Work}\label{sec:review}

A comprehensive review of coverage path planning methods can be found in~\cite{shen2026coverage_survey}. In this section, we briefly discuss representative online CPP methods.

Gabriely and Rimon~\cite{gabriely2001spanning} proposed the spanning tree covering algorithm, which constructs a spanning tree over a coarse grid and generates the coverage path by traversing around the tree. Each coarse cell is divided into four fine cells according to the robot footprint. Full spiral STC~\cite{gabriely2003competitive} further incorporates partially occupied cells into an augmented tree to improve coverage resolution. Gonzalez et al.~\cite{gonzalez2005bsa} developed the backtracking spiral algorithm, which generates spiral-like coverage motions by following obstacles and covered regions along a fixed lateral direction. When no local uncovered cell is available, the robot backtracks to resume coverage elsewhere. Ferranti et al.~\cite{ferranti2007brick} proposed the brick-and-mortar algorithm, which expands inaccessible regions while preserving traversability among the remaining accessible cells. Luo and Yang~\cite{luo2008bioinspired} developed a biologically inspired neural-network-based coverage method in which each grid cell is represented by a neuron. Uncovered cells provide excitatory inputs while obstacles provide inhibitory inputs. The robot follows the resulting neural activity landscape, whose propagation can also guide it toward distant uncovered regions when no local target is available.

Viet et al.~\cite{viet2013ba} proposed the BA$^*$ algorithm, which selects the next uncovered cell according to a predefined directional priority and consequently produces back-and-forth motions. When the local sweep can no longer continue, the robot moves to a previously identified backtracking location and resumes coverage from the remaining uncovered region. Song and Gupta~\cite{song2018} proposed the $\varepsilon^*$ algorithm, which employs multiscale adaptive potential surfaces for coverage decision-making. The robot normally selects targets using the lowest-level potential surface in its local neighborhood, while higher-level potential surfaces are used when the robot encounters dead-end situations. Hassan and Liu~\cite{hassan2019ppcpp} proposed the predator-prey coverage path planning algorithm, which evaluates neighboring cells using a multi-objective reward function. The reward considers the distance from a virtual stationary predator, motion smoothness, and the tendency to follow the uncovered-region boundary. The candidate with the highest reward is selected as the next coverage target.

Shen et al.~\cite{shen2025_cap} proposed the connectivity-aware hierarchical coverage path planning algorithm, which dynamically identifies disconnected uncovered subareas and represents them using a global graph. Collision-free connections between subareas are used to compute a TSP-based traversal tour. At the local level, the algorithm switches between exploratory coverage in partially observed subareas and optimized coverage paths in fully observed regions. Shen et al.~\cite{shen2026} proposed C$^*$, a sampling-based online CPP algorithm that incrementally constructs a sparse rapidly covering graph (RCG) from sensor observations. The RCG provides non-local guidance for coverage waypoint selection and dead-end escape. C$^*$ also detects potential coverage holes during navigation and generates locally optimized trajectories to cover them before continuing the main coverage process.

Some CPP methods have also been developed for 3D structure coverage~\cite{shen2022ct,feng2024fc}, multi-robot systems~\cite{shen2025multi,wang2025mac}, curvature-constrained robots~\cite{shen2019online,maini2022online}, tethered robots~\cite{peng2025spanning,shnaps2014online}, and energy-constrained robots~\cite{shen2020,dogru2022eco}.

\section{Proposed Algorithm}\label{sec:algorithm}
 
Let $\mathcal{A}\subset\mathbb{R}^{2}$ denote the initially unknown search area and is discretized into a set of cells $\mathcal{T}=\{\tau_j\}_{j=1}^{|\mathcal{T}|}$. During online operation, the robot updates the state of each cell using onboard sensor measurements. Accordingly, the tiling is partitioned into obstacle, free, and unknown cells, i.e., $\mathcal{T}=\mathcal{T}_o\cup \mathcal{T}_f\cup \mathcal{T}_u$, where the free cells are further divided into covered and uncovered subsets, $\mathcal{T}_f=\mathcal{T}_f^c\cup \mathcal{T}_f^u$. The proposed method consists of three components: online space decomposition, global tour planning, and local coverage planning. The key idea is to retain the hierarchical decomposition structure that naturally emerges during online coverage and directly exploit this structure for global task sequencing.

\subsection{Online Space Decomposition}
\label{detectSubarea}

At the beginning of the coverage task, the entire search area $\mathcal{A}$ is represented by the root node $n_0$ of a decomposition tree $\mathcal{G}=(\mathcal{N},\mathcal{E})$, where $\mathcal{N}$ denotes the set of subarea nodes and $\mathcal{E}$ represents the parent-child relationships generated during online decomposition. The root node is set as target $n_{\mathrm{target}}$. The robot performs local coverage in the target subarea $\mathcal{A}_{\mathrm{target}}$. As coverage progresses, newly observed obstacles and covered cells may partition the remaining uncovered portion of this region into multiple disconnected subareas. Theses subareas are identified through recursive search over neighboring uncovered cells. Starting from an unlabeled cell in target subarea, all connected neighboring uncovered cells are recursively assigned the same subarea index. The procedure is repeated until all uncovered cells belonging to the current region have been assigned. 

Suppose that $\mathcal{A}_{\mathrm{target}}$ is divided into $L$ disconnected components, i.e., $\mathcal{A}_{\mathrm{target}}\rightarrow
\{\mathcal{A}_{1},\mathcal{A}_{2},\ldots,\mathcal{A}_{L}\}$. The corresponding node $n_{\mathrm{target}}$ is then expanded by inserting $\{n_{1},n_{2},\ldots,n_{L}\}$ as its child nodes in $\mathcal{G}$, where each child represents one newly identified subarea. The tree explicitly records how the search area evolves from coarse regions into progressively smaller coverage subareas. Each active subarea is further assigned one of two states. A subarea is classified as exploring if it contains at least one uncovered free cell adjacent to the unknown region $T_u$. Otherwise, it is classified as explored. Only leaf nodes of the tree are considered active coverage tasks. As the robot explores the environment, the decomposition tree is incrementally expanded until all active leaves have been completely covered.

\subsection{Global Coverage Tour Planning}
\label{globalCover}

A global coverage tour $\gamma_{global}=\{n_{(1)},n_{(2)},\ldots,n_{(M)}\}$ containing the active subareas that remain to be covered is constantly maintained and updated online. Given the child nodes $\{n_{1},n_{2},\ldots,n_{L}\}$ of target node $n_{\mathrm{target}}$. The child nodes are first ordered according to their coverage states. Explored subareas receive higher priority than exploring subareas. Within the same state category, the children are ranked according to their transition distance from the robot. Let $p_R$ denote the current robot position and $p_{i}$ the representative position of subarea $\mathcal{A}_{i}$. The transition cost is defined as $d(p_R,p_{i})$, where $d(\cdot,\cdot)$ is the Euclidean distance. The children are therefore sorted lexicographically such that an explored child is always preferred over an exploring child, while children having the same state are ordered by increasing transition distance.

Let $\Gamma=\{n_{\alpha_1},n_{\alpha_2},\ldots,n_{\alpha_L}\}
$ denote the resulting ordered child sequence. The global tour is updated by replacing the current target node  $n_{\mathrm{target}}$ with $\Gamma$. For example, let $\{n_1,n_2,n_3\}$ be the current global tour $\gamma_{global}$. Let $n_1$ be the current target node $n_{\mathrm{target}}$. Suppose $n_1$ is decomposed into $n_{11}$ and $n_{12}$, then the new tour is $\{n_{11},n_{12},n_2,n_3\}$ and $n_{11}$ is set as target. If $n_{11}$ is later decomposed into $n_{111}$ and $n_{112}$, the sequence becomes $\{n_{111},n_{112},n_{12},n_2,n_3\}$. Thus, newly generated descendants of the current subarea are prioritized before the planner switches to other branches. The resulting behavior follows the hierarchical structure revealed during online coverage while avoiding repeated global optimization over all remaining subareas. If the current target is completely covered without further decomposition, it is simply removed from the tour, i.e., $
\gamma_{\mathrm{global}}
\leftarrow
\gamma_{\mathrm{global}}
\setminus
n_{\mathrm{target}},
$ and the first remaining node becomes the next target node.

\subsection{Local Coverage Path Planning} \label{localCover}

 The local planner generates the coverage motion inside the corresponding subarea of target node $n_{\mathrm{target}}$. 

For an exploring subarea, the robot follows a greedy strategy that selects an uncovered cell $\tau_{\mathrm{target}}$ from its local neighborhood according to a predefined directional priority. The selected cell is marked as covered once visited, producing a back-and-forth coverage pattern while simultaneously discovering new environmental information. As the robot continues covering an exploring subarea, the updated map may reveal that the remaining uncovered region has been divided into multiple disconnected components. In this case, the current node is expanded in the decomposition tree and the global tour is updated using the procedure described in Sections~\ref{detectSubarea} and~\ref{globalCover}. For an explored subarea, its geometry is already completely known. The local planner therefore generates a shortest coverage path $
\gamma_{\mathrm{local}}
$ that starts from the current robot position $p_R$, visits all remaining uncovered cells in the selected subarea, and terminates at the designated target cell.

The global and local planning steps are repeatedly executed until $\gamma_{\mathrm{global}}=\varnothing$, which indicates that no uncovered subarea remains and complete coverage has been achieved.

\RestyleAlgo{ruled}
\LinesNumbered
\begin{algorithm}[t]
\footnotesize

$\mathcal{G} \leftarrow n_0$; 
$\gamma_{global} \leftarrow \{n_0\}$; 
$n_{target} \leftarrow n_0$\;

\While{$\gamma_{global} \neq \emptyset$}{

    $\left\{\mathcal{T},p_{\mathcal{R}}\right\}
    \leftarrow \textbf{UpdateSensorData}()$\;

    \If{\textbf{isExploring}($n_{target}$)}
    {
        $\{\mathcal{A}_{\ell}\}_{\ell=1}^{L}
        \leftarrow
        \textbf{IdentifySubarea}(n_{target})$\;

        \If{$L>1$}
        {
            $\mathcal{G}
            \leftarrow
            \textbf{UpdateTree}
            (\mathcal{G},n_{target},
            \{\mathcal{A}_{\ell}\}_{\ell=1}^{L})$\;

            $\Gamma
            \leftarrow
            \textbf{OrderChildren}
            (\{\mathcal{A}_{\ell}\}_{\ell=1}^{L},
            p_{\mathcal{R}})$\;

            $\gamma_{global}
            \leftarrow
            \textbf{UpdateGlobalTour}
            (\gamma_{global},n_{target},\Gamma)$\;

            $n_{target}
            \leftarrow
            \gamma_{global}.\text{top}()$\;
        }
    }

    \If{$\textbf{isComplete}(n_{target})$}
    {
        $\gamma_{global}
        \leftarrow
        \gamma_{global}-n_{target}$\;

        \If{$\gamma_{global}\neq\emptyset$}
        {
            $n_{target}
            \leftarrow
            \gamma_{global}.\text{top}()$\;
        }
    }

    \textbf{LocalCover}($n_{target}$)\;
}

\caption{Proposed Algorithm}
\label{alg:proposed}
\end{algorithm}
\setlength{\textfloatsep}{0pt}

\subsection{Time Complexity Analysis}
The computational complexity of the proposed method is evaluated for the main operations of online space decomposition, global tour planning, and local coverage planning. 

During online space decomposition, the robot recursively searches the uncovered free cells to identify disconnected subareas. Since each relevant uncovered cell is labeled at most once, this operation has a worst-case complexity of $O(|\mathcal{T}_f^u|)$, where $\mathcal{T}_f^u$ denotes the set of currently uncovered free cells. Suppose that the current target subarea is decomposed into $L$ child subareas. Updating the decomposition tree requires only inserting the newly generated child nodes and their parent-child connections, resulting in a complexity of $O(L)$. For global tour planning, the newly generated children are first classified according to their explored or exploring states. Within each category, their priorities are determined using the Euclidean distances between the robot and the representative positions of the child subareas. Distance evaluation requires $O(L)$ operations, while sorting the children requires $O(L\log L)$ time.  Once the next target subarea is selected, a shortest collision-free transition path from the current robot position to the target is computed using the A* algorithm. In the worst case, the search is performed over the currently known free cells, resulting in a complexity of $O(|\mathcal{T}_f|^2)$, where $\mathcal{T}_f$ denotes the set of currently known free cells. For local coverage planning, an exploring subarea is covered using a greedy strategy with $O(1)$ complexity for each local decision. For an explored subarea, the local planner has a worst-case complexity of $O(|\mathcal{A}_{\mathrm{target}}|^2)$, where $|\mathcal{n}_{\mathrm{target}}|$ denotes the number of cells of the target subarea.

Therefore, the overall worst-case complexity of one planning update is $O\left(
|\mathcal{T}_f^u|
+
L\log L
+
|\mathcal{T}_f|^2
+
|\mathcal{A}_{\mathrm{target}}|^2
\right)$. Since $
|\mathcal{T}_f|\geq |\mathcal{T}_f^u|
$ and $|T_f|\geq |\mathcal{A}_{\mathrm{target}}|$, and the number of child subareas $L$ generated by a single decomposition event is generally much smaller than $|\mathcal{T}_f|$, the overall time complexity is dominated by the shortest-path search and can be expressed as $O(|\mathcal{T}_f|^2)$.

\begin{figure*}[t]
    \centering
    \subfloat[Coverage paths generated by different algorithms in scene 1.]{
    \includegraphics[width=0.85\textwidth]{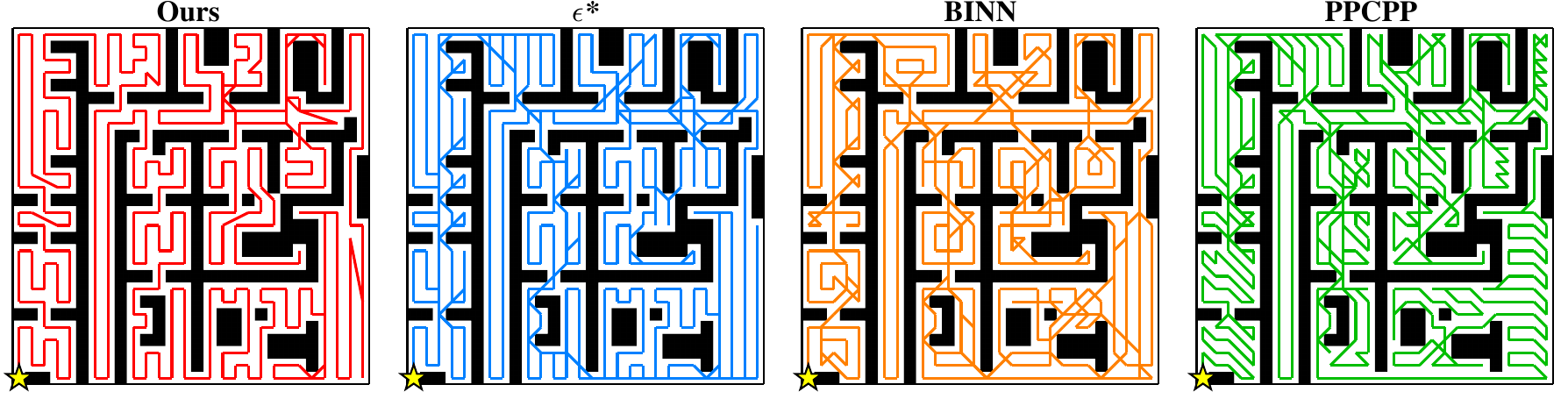}\label{fig:path_simulation_scenario1}}\vspace{1em}\\ 
    \centering
    \subfloat[Coverage paths generated by different algorithms in scene 2.]{
    \includegraphics[width=0.85\textwidth]{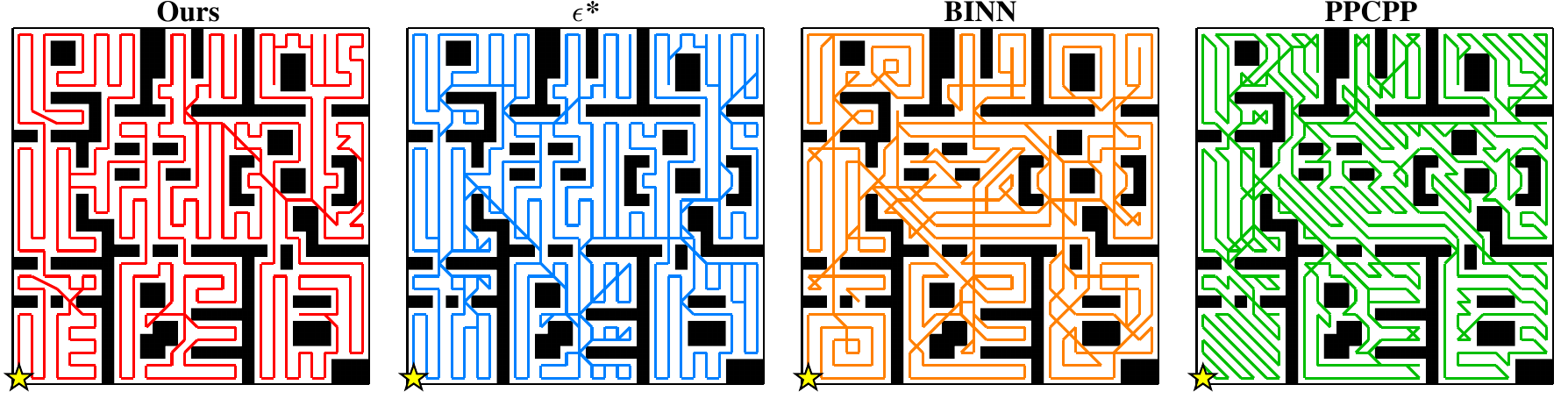}\label{fig:path_simulation_scenario2}}\vspace{1em}\\
    \centering
    \subfloat[Comparison of performance metrics.]{
    \includegraphics[width=0.6\textwidth]{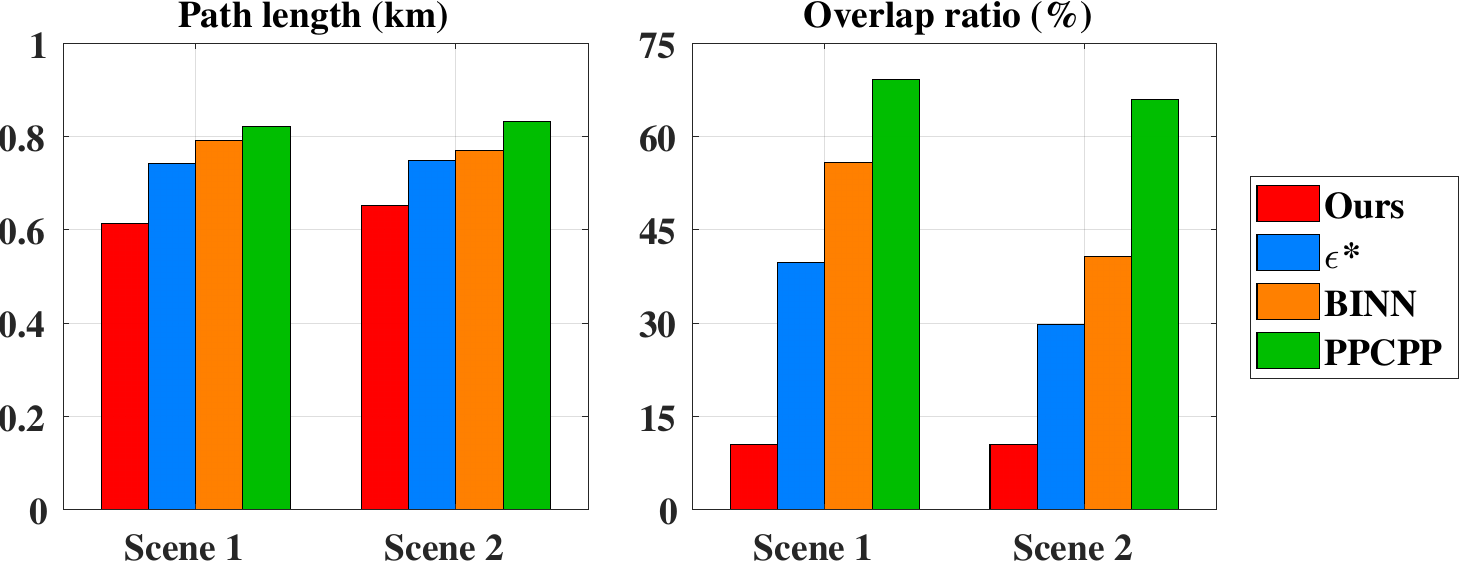}\label{fig:metric_simulation}}\\
    \caption{Performance comparison of proposed algorithm with the baseline algorithms in simulations.}
  \label{fig:result_simulation}
\end{figure*}
\section{Results and Discussion}
\label{sec:results}
The performance of proposed algorithm is evaluated by high-fidelity simulations on Gazebo. As shown in Fig.~\ref{fig:result_simulation}, four complex scenes of dimensions $90 m \times 90 m$ are generated with different obstacle layouts and partitioned into a $30 \times 30$ tiling structure for mapping and coverage. Each scene depicts a real situation (e.g., office). A car-like robot is simulated with a maximum speed of $1m/s$ and initialized at the bottom-left corner of the space. It is equipped with a lidar with a detection range of $8m$. Three baseline algorithms including  $\epsilon^*$~\cite{song2018}, BINN~\cite{luo2008bioinspired}, and PPCPP~\cite{hassan2019ppcpp} are selected for performance comparison in terms of path length and overlap ratio. Figs.~\ref{fig:path_simulation_scenario1}-\ref{fig:path_simulation_scenario2} show the coverage paths generated by different algorithms in two scenes. As seen, the proposed algorithm provides complete coverage with less overlapping paths. This is because it captures the information of disconnected subareas and and then computes a global tour to guide the coverage process, which helps improve overall coverage efficiency. Additionally, it performs TSP-based optimal coverage path within the explored subareas to further minimize overlapping paths and local coverage time. In contrast, the baseline algorithms follow greedy strategies to achieve complete coverage without a
global perspective of the entire search area, thus leading to highly overlapping paths and sub-optimal motions. Fig.~\ref{fig:metric_simulation} provides quantitative comparison results in terms of path length and overlap ratio. Overall, the proposed algorithm achieves significant improvements over the baseline algorithms in all metrics.

\section{Conclusions and Future Work} \label{sec:conclusions}

This paper presents a hierarchical coverage path planning method for efficient online coverage of unknown environments. During navigation, the search space is progressively decomposed into disconnected subareas, which are organized in an incrementally constructed decomposition tree that preserves their parent-child relationships. The tree structure is directly exploited for global tour planning, where newly generated child subareas are prioritized according to their exploration states and distances from the robot. Combined with adaptive local coverage planning, the proposed method enables the robot to continuously update its coverage trajectory as new environmental information becomes available. Experimental evaluations in simulation environments demonstrate the effectiveness of the proposed method in terms of path length and overlap ratio. Future work will investigate extensions of the proposed framework to multi-robot coverage, where the decomposition tree can be further exploited for online task allocation and coordination among robots. Another direction is to extend the method to dynamic environments~\cite{shen2023smart,shen2026motion,shen2026learning}, where the decomposition structure needs to adapt to time-varying obstacles.


\balance
\bibliographystyle{IEEEtran}
\bibliography{reference}

\end{document}